\documentclass{article}
\usepackage{ijcai26}

\usepackage{times}
\usepackage{soul}
\usepackage{url}
\usepackage[hidelinks]{hyperref}
\usepackage[utf8]{inputenc}
\usepackage[small]{caption}
\usepackage{graphicx}
\usepackage{amsmath}
\usepackage{amsthm}
\usepackage{amssymb}
\usepackage{booktabs}
\usepackage{algorithm}
\usepackage{algorithmic}
\usepackage[switch]{lineno}

\title{A Unified Knowledge Embedded Reinforcement Learning-based Framework for Generalized Capacitated Vehicle Routing Problems}

\author{
Wen Wang$^1$\and
Xiangchen Wu$^1$\and
Liang Wang$^{1}$\and
Hao Hu$^1$\And
Xianping Tao$^1$
\affiliations
$^1$ Nanjing University
\emails
\{wangwen, ixc\}@smail.nju.edu.cn,
\{wl, myou, txp\}@nju.edu.cn
}

\begin{document}

\maketitle

\begin{abstract}
The Capacitated Vehicle Routing Problem (CVRP) is a fundamental NP-hard problem with broad applications in logistics and transportation. Real-world CVRPs often involve diverse objectives and complex constraints, such as time windows or backhaul requirements, motivating the development of a unified solution framework.
Recent reinforcement learning (RL) approaches have shown promise in combinatorial optimization, yet they rely on end-to-end learning and lack explicit problem-solving knowledge, limiting solution quality.
In this paper, we propose a knowledge-embedded framework inspired by the Route-First Cluster-Second heuristics. 
It incorporates knowledge at two levels: (1) decomposing CVRPs into the route-first and cluster-second subproblems, and (2) leveraging dynamic programming to solve the second subproblem, whose results guide the RL-based constructive solver to solve the first problem.
To mitigate partial observability caused by problem decomposition, we introduce a unified history-enhanced context processing module. 
Extensive experiments show that this framework achieves superior solution quality compared with state-of-the-art learning-based methods, with a smaller gap to classical heuristics, demonstrating strong generalization across diverse CVRP variants.
\end{abstract}

\section{Introduction}

The Capacitated Vehicle Routing Problem (CVRP) is a fundamental research topic in combinatorial optimization and serves as a canonical model for numerous real-world applications, including urban logistics \cite{jeong2025optimizing}, last-mile delivery \cite{tiwari2023optimization}, postal distribution \cite{sbai2022two}, and transportation planning \cite{leng2021distribution}, where goods must be delivered from a central depot to a set of geographically distributed customers.
In practical applications, CVRP instances often incorporate additional constraints to reflect real-world operational requirements.
For example, time window constraints \cite{vidal2014unified} enforce that customers must be served within specified intervals, while open route constraints \cite{li2007open} allow vehicles to finish their service without returning to the depot.
Given the NP-hard nature \cite{lenstra1981complexity} and the increased complexity introduced by these variants, developing a unified and efficient approximate solver has become an important and long-standing research objective.

With the development of learning-based methods in the field of combinatorial optimization, a number of foundation models for VRPs have recently been proposed \cite{berto2024routefinder,Drakulic2025GOAL,zhou2024mvmoe}.
These methods introduce advanced architectural designs inspired by Large Language Models (LLMs), incorporating components such as multi-head mixed-attention blocks, transformer-based encoders, and mixture-of-experts (MoE) structures to construct a unified network architecture capable of addressing diverse VRP variants.
From a learning perspective, these models typically adopt POMO-style \cite{kwon2020pomo} reinforcement learning or imitation learning paradigms to train neural networks that directly generate high-quality routing solutions in an end-to-end manner.
In this solving framework, the incorporation of domain knowledge is manifested in two ways: (1) the design of network architectures guided by problem-specific insights; and (2) solution samples generated by classical solvers.
Although promising results have been achieved, relying solely on model- or data-driven domain knowledge overlooks the importance of algorithmic design in solving classical problems, thereby limiting further improvements in solution quality.

To directly leverage domain knowledge in solving CVRPs, we exploit the decomposable nature of the problem.
Problem decomposition has a long-standing history in the study of CVRPs, particularly within the field of evolutionary computation \cite{zhang2022review}.
Among these approaches, the Route-first Cluster-second (RFCS) heuristic \cite{Prins2014OrderfirstSM} represents one of the most influential strategies, and numerous variants have been developed following this principle.
Specifically, the RFCS heuristic first constructs a global route covering all customer nodes (excluding the depot) using heuristic methods such as TSP solvers.
It then applies a polynomial-time optimal algorithm to divide the global route into feasible vehicle tours, thereby producing a valid solution to original CVRP variants.
Manually designed algorithms can naturally incorporate various operational constraints.
More importantly, once the global route is generated, the subsequent process typically runs in polynomial time, ensuring high computational efficiency.
For example, \cite{Vidal2015TechnicalNS} give a $O(n)$ algorithm given a global route for minsum CVRP, and \cite{wang2025solving} gives a binary search-based splitting algorithm for minmax VRPs.
Nevertheless, such methods generally rely on TSP-based heuristics to construct the global route, and the optimal TSP tour does not necessarily yield an optimal solution to the original CVRP variants after a cluster-second operation.
In this paper, we introduce reinforcement learning to automatically learn the route first part, avoiding the limitations of handcrafted TSP-based heuristics.

However, the problem decomposition mechanism introduces a partial observability challenge: the context of a partial route cannot be fully assessed until the entire route is generated and the subsequent splitting stage is completed.
To address this issue, we introduce a history-enhanced module as the core of our context processor.
This module use the LSTM \cite{Hochreiter1997LongSM} architecture to maintain a temporal hidden state that continuously aggregates historical information throughout the route construction process, allowing the model to infer the latent information of the problem even when the final evaluation signal is postponed.
Beyond mitigating partial observability, the history-enhanced context representation also provides a unified neural architecture across different VRP variants.
By encoding the evolving decision context into a learned hidden state, the framework eliminates the need for manually designed problem-specific contexts.
As a result, the same network architecture can generalize seamlessly across multiple problem types—such as open-route, backhaul-request, and time-window CVRPs—without additional adaptation.

The main contributions can be summarized as follows.

\begin{itemize}
    \item We propose a unified \textbf{knowledge-embedded RFCS framework} that integrates RL with the route-first cluster-second (RFCS) heuristic to address a wide range of CVRP variants.
    This framework combines general learning-based methods and algorithmic design principles, enabling flexible adaptation to various types of problem-specific constraints.
    
    \item We introduce a \textbf{history-enhanced context processing} module that preserves a unified neural network architecture across general CVRP variants. 
    This module simplifies the design of context features by exploiting a common representation provided by the route-first component, enabling effective handling of complex constraints.
    
    \item We perform extensive experiments to evaluate the solution quality and transfer capabilities of the framework across diverse CVRP settings. 
    The results demonstrate that the proposed knowledge-embedded RFCS framework consistently outperforms existing learning-based methods and exhibits strong zero-shot generalization to unseen problem variants.

\end{itemize}

\section{Related Work}

Since the introduction of neural networks into combinatorial optimization by the Pointer Network \cite{vinyals2015pointer}, a series of learning-based combinatorial optimization solvers have subsequently emerged, collectively referred to as Neural Combinatorial Optimization (NCO).
From the perspective of solution paradigms, learning-based approaches can be broadly classified into three categories \cite{wu2024neural}: (1) Learning to Construct (L2C), which focuses on directly generating solutions; (2) Learning to Improve (L2I), which aims to refine or enhance existing solutions; and (3) Learning to Predict (L2P), which leverages neural networks to support and augment traditional operations research methods.
In the L2C paradigm, the Attention Model (AM) \cite{koolattention} proposed a Transformer-based encoder-decoder architecture to directly generate solutions. Subsequently, POMO \cite{kwon2020pomo} and Sym-NCO \cite{kim2022sym} introduced multi-start and symmetry-based \cite{shi2025symmetry,yu2024leveraging} training strategies, respectively, during the solution generation process, further enhancing solution quality.
More recent studies have focused on improving generalization capability \cite{gao2024towards}, addressing diverse objectives and constraints \cite{son2024equity,zheng2024dpn,bi2024learning}, and scaling up to larger problem instances \cite{luo2023neural,kim2022scale}.
This paradigm achieves a favorable balance between solution quality and inference efficiency, and it is also adopted in this work.
The L2I paradigm originated from learning local rewriting operations \cite{chen2019learning}, and was later extended to the learning of heuristic operators such as 2-OPT and cross-exchange \cite{d2020learning,hottung2020neural,sui2021learning,wu2021learning}. In recent years, research in this direction has increasingly focused on large-scale problem solving \cite{cheng2023select,li2025destroy}.
Although this paradigm is more suitable for handling large-scale problems, it is relatively difficult to employ a unified framework to address diverse types of constraints.
The L2P paradigm integrates traditional algorithmic approaches with learning-based methods, such as combining learning with the LKH algorithm \cite{xin2021neurolkh,zheng2021combining}, integrating learning with dynamic programming \cite{kool2022deep,xu2020deep}, and leveraging large language models (LLMs) and meta-learning for automated algorithm design \cite{liu2024evolution,dernedde2025moco}.
Methods of this type, by incorporating principles of algorithm design, have the potential to improve solution quality. 
In this work, we adopt a similar approach within the L2C framework, introducing knowledge as a reward signal, which facilitates the handling of general forms of constraints.
\section{Preliminaries}\label{sec:problem}

In this section, we first introduce the general form CVRPs.
Then, we summarize the RL formulation for CVRPs, describing how solutions can be modeled as sequential actions and optimized under the RL paradigm.

\subsection{General Form CVRPs}

The CVRPs can be defined on a complete graph $\mathcal{G} = (\mathcal{V}, \mathcal{E})$, 
where $\mathcal{V} = \{v_0, v_1, \cdots, v_n\}$ is the set of nodes, with $v_0$ representing the depot and $\{v_1, \cdots, v_n\}$ representing the customer nodes.
Each edge $(i, j) \in \mathcal{E}$ is associated with a non-negative cost $c_{ij}$.
As a common practice, each node is assigned a two-dimensional coordinate, and the Euclidean distance is used as the cost.
Vehicles with identical capacity $Q$ are stationed at the depot.
Each customer node $i \in \{1, \cdots, n \}$ has a nonzero demand $q_i \le Q$ that must be satisfied by exactly one vehicle.
Each vehicle travels from the depot node, serves customers and returns to the depot.
The total demand of the served nodes should not exceed the capacity.

Following \cite{berto2024routefinder}, we consider a generalized formulation of the CVRP that incorporates multiple additional constraints:
(1) Open routes --- vehicles are not required to return to the depot;
(2) Backhaul demands --- customer requests may involve both delivery and pickup operations;
(3) Distance limits --- the travel distance of each route must not exceed a predefined threshold; and
(4) Time windows — each customer must be served within a specified time interval.
Each additional constraint can be independently activated or deactivated, resulting in sixteen problem variants, denoted as CVRP, OVRP, VRPB, VRPL, VRPTW, etc.

We aim to minimize the total travel cost over all vehicle routes.
Let $\{\mathcal{R}_k\}_{k=1}^K$ denote a feasible routing plan, where $\mathcal{R}_k$ is the sequence of nodes served by vehicle $k$.
The cost of route $\mathcal{R}_k$ is given by
\begin{equation}\label{eq:vehicle_cost_or}
C_k = \sum_{(i,j)\in \mathcal{R}_k} c_{i,j},
\end{equation}
with $c_{i,j}$ representing the travel cost between nodes $i$ and $j$.
The min-sum objective is then formulated as
\begin{equation}\label{eq:minsum_objective_or}
\min_{\{\mathcal{R}_k\}_{k=1}^{K}} \; \sum_{k=1}^{K} C_k,
\end{equation}
subject to the following feasibility conditions:
1) each customer is visited exactly once; 2) the total demand on each route does not exceed vehicle capacity; and 3) all additional constraints are satisfied.
Here, $K$ denotes the number of active vehicles in the solution.

\subsection{Reinforcement Learning for CVRPs}

In the RL framework for CVRPs, a constructive solver sequentially builds a solution by selecting one customer node at a time.  
Formally, at step $t$, the solver selects an action $a_t \in V_t$ based on a context $c_t$, which corresponds to visiting a customer node next, according to a stochastic policy
$
a_t \sim \pi_\theta(a_t \mid c_t),
$
parameterized by neural network parameters $\theta$.
After visiting node $a_t$, the state is updated to $c_{t+1}$, and the set of available actions is reduced: $V_{t+1} = V_t \setminus \{a_t\}$.

Once all customers are visited and feasible routes are constructed, the solver receives a reward $R$ based on the objective.
The RL training objective is to maximize the expected reward over all possible construction sequences:
\begin{equation}\label{eq:rl_obj}
   J(\theta) = \mathbb{E}_{\tau \sim \pi_\theta} \left[ R(\tau) \right],
\end{equation}
where $\tau = (a_0, a_1, \cdots, a_T)$ denotes a trajectory generated by the policy.  
Gradient-based optimization can be performed using policy gradient methods, e.g., REINFORCE \cite{williams2025}, optionally enhanced with baselines to reduce variance:
\begin{equation}\label{eq:reinforce_objective}
\nabla_\theta J(\theta) = \mathbb{E}_{\tau \sim \pi_\theta} \Big[ \sum_{t=0}^{T} \nabla_\theta \log \pi_\theta(a_t \mid c_t) \, (R(\tau) - b) \Big],    
\end{equation}

where \(b\) is a baseline value.

\section{The proposed method}

\begin{figure*}[htb]
    \centering
    \includegraphics[width=0.8\linewidth]{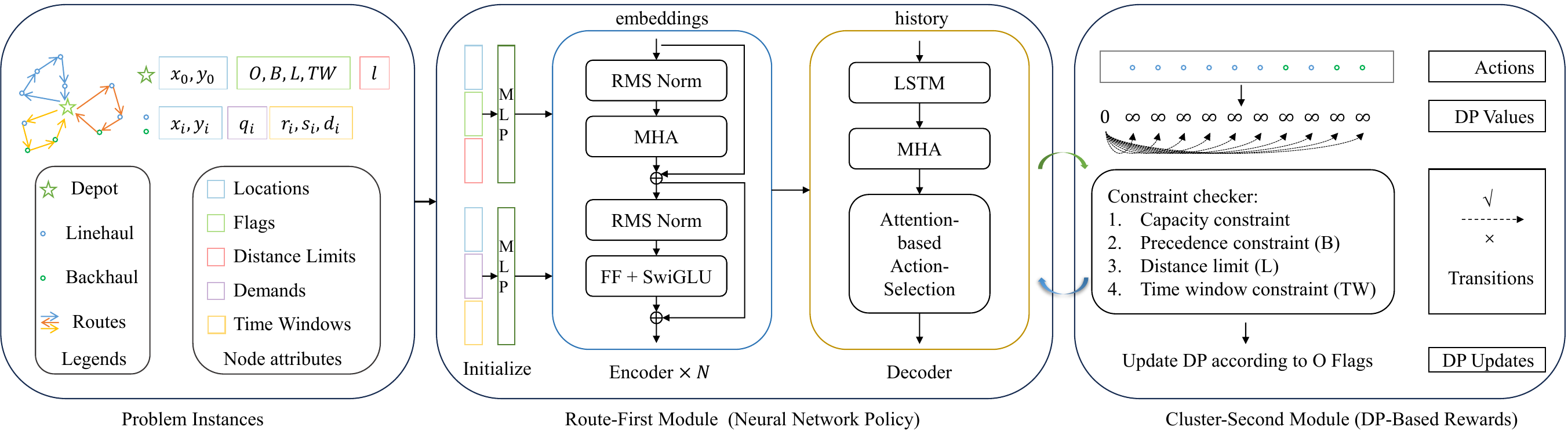}
    \vspace{-6 pt}
    \caption{The Overall Framework of the Proposed Knowledge-embedded RFCS method.}\label{fig:framework}
\end{figure*}

Unlike a fully end-to-end learning paradigm, we propose a knowledge-embedded RL framework for solving generalized CVRPs.
The framework integrates problem-solving principles with algorithmic design insights into the training process to improve solution quality.
It also simplifies transition modeling in the training environment, reduces the complexity of incorporating new constraints, and enables a unified network architecture applicable across multiple CVRP variants.

\subsection{The Overall Framework}\label{sec:framework}

The overall framework is illustrated in Figure~\ref{fig:framework}.
We adopt a \textbf{route-first, cluster-second} heuristic structure.
The RL component focuses on the route-first stage, where the objective is to generate a sequence that visits all customer nodes exactly once, resulting in a permutation of customer indices.
All complex constraints are handled by a dynamic programming-based cluster-second module.
Given any customer permutation, this module determines the optimal splitting of the sequence into feasible vehicle routes while satisfying additional constraints such as time windows and duration limits.

Although the framework consists of two components, the reward signal is computed based on the solution produced by the cluster-second module.
This reward is then used to optimize the route-first policy through RL, thereby aligning the optimization objective of the RL model with that of the original problem formulation.
In this way, the framework eliminates the dependence on heuristic TSP solvers and achieves a direct, learning-based formulation of the routing problem.

\subsection{The Unified Environment}

The framework allows problems with heterogeneous constraints or objectives to be reduced to an identical learning structure, thus we adopt a unified formulation for constructing the RL environments under different constraint settings, as described below.

\textbf{States:} At step $t$, the environment maintains a state or context $s_t$ that summarizes all relevant information needed for decision-making:

\begin{equation}\label{eq:state_context}
    s_t = \{ \mathbf{X}, \mathbf{q}, \mathbf{TW}, l, n_0, n_1, v_t, \mathbf{m}, \mathbf{f} \},
\end{equation}

where:
\begin{itemize}
    \item \(\mathbf{X} \in \mathbb{R}^{n \times 2}\) denotes the coordinates of depot and customer nodes. We use $x_i, y_i$ for the $i$-th node;
    \item \(\mathbf{q} \in \mathbb{R}^{n \times 1}\) represents the customer demands, including negative values for pickup nodes. We use $q_i$ for the $i$-th node;
    \item \(\mathbf{TW} \in \mathbb{R}^{n \times 3}\) encodes time windows and service durations. We use $r_i, s_i, d_i$ to denote the start time, the service duration, and the deadline for node $i$.
    \item \(l \in \mathbb{R}\) represents distance limits;
    \item \(n_0, n_1 \in \mathbb{R}\) track the cumulative delivered and picked-up quantities for each route;  
    \item \(v_t \in \{0,\dots,n\}\) is the current node being visited;  
    \item \(\mathbf{m} \in \{0,1\}^{n}\) is the action mask, indicating feasible next visited nodes;  
    \item \(\mathbf{f} \in \{0,1\}^{4}\) encodes the variant features (O, B, L, TW) of the current instance, where each element represents a binary flag indicating whether the corresponding constraint is active.
\end{itemize}

When certain attributes are absent, default values are used as placeholders to maintain a unified data structure.

\textbf{Actions:} The action at step \(t\) is selecting the next node \(a_t \in \{1, \dots, n + 1\}\) to visit. 

\textbf{Transitions:} Benefiting from the proposed design, state updates can be performed through a simple and efficient procedure.
The variables $n_0, n_1, v_t$ are updated directly according to their definitions, while $\mathbf{m}$ is refreshed by setting the value of each visited node to zero.
The responsibility for checking action feasibility is fully delegated to the cluster-second module.

\textbf{Rewards:} We employ a dynamic programming-based solver (Section~\ref{sec:dp_rewards}) to optimally convert the solutions to the original problem.
The reward is defined as the negative of the total cost computed by this solver.

We hypothesize that this structural consistency contributes substantially to the effectiveness and generality of the proposed framework.

\subsection{Dynamic Programming Solver}\label{sec:dp_rewards}

To compute the optimal cost of the route-first sequence, we develop a dynamic programming (DP)-based approach capable of naturally handling diverse constraint types.
Let a given sequence of customer nodes (excluding the depot) be denoted as \(\pi = (a_1, a_2, \dots, a_n)\), and let \(p[i]\) represent the minimum total cost to serve the first \(i\) nodes.  
The DP formulation recursively computes
\begin{equation}\label{eq:dp}
    p[i] = \min_{j < i} \{ p[j] + \overline{C}(j+1,i) \},
\end{equation}
where $\overline{C}(j+1,i)$ is the cost of serving nodes $(\pi_{j+1}, \dots, \pi_i)$ as a single vehicle route, subject to capacity, pickup-and-delivery precedence, distance, and time window constraints.
To naturally handle these constraints, a simple yet effective strategy is adopted: if the sequence $(\pi_{j+1}, \dots, \pi_i)$ cannot form a feasible single vehicle route, $\overline{C}(j+1, i)$ is defined as $+\infty$. 
Algorithm~\ref{alg:msdp} details the procedure.
Specifically, lines 5-6 manage the backhaul-related variants, and all other constraints, including capacity, time windows, and duration limits, are handled through their respective conditional checks.
It is straightforward to extend this algorithm to incorporate additional constraints.
For example, adding support for the mixed-backhaul constraint \cite{zhou2024mvmoe} can be achieved by simply adjusting the condition in line 6.

\begin{algorithm}[!t]
\caption{Dynamic Programming-based Solver}
\label{alg:msdp}
\begin{algorithmic}[1]

\REQUIRE Customer sequence $\pi = (a_1, a_2, \cdots, a_n)$, node attributes, variant flags
\ENSURE Minimum total route cost $p[n]$

\STATE Initialize $p[0] \gets 0$, $p[i] \gets +\infty$ for all $i = 1, \dots, n$
\FOR{$t = 0$ \textbf{to} $n-1$}
    \STATE Initialize $l_o \gets 0$, $l_i \gets 0$, $t_{now} \gets 0$, \textit{cost} $\gets 0$
    \FOR{$i = t+1$ \textbf{to} $n$}
        \STATE $l_o \gets l_o + \max(q_{a_i}, 0)$,\quad $l_i \gets l_i + \max(-q_{a_i}, 0)$

        \IF{$l_o > Q$ \OR $l_i > Q$ \OR precedence violated}
            \STATE \textbf{break}
        \ENDIF
        \STATE $\textit{cost} \gets \textit{cost} + c_{a_{i-1}, a_i}$
        \IF{TW active}
            \STATE $t_{now} \gets \max(t_{now}, r_{a_i}) + s_{a_i} $
            \IF{$t_{now} > d_{a_i}$}
                \STATE \textbf{break}
            \ENDIF
        \ENDIF
        \IF{L active \AND $\textit{cost} > l$}
            \STATE \textbf{break}
        \ENDIF
        \IF{\textit{O} active}
            \STATE $p[i] \gets \min(p[i], \; p[t] + \textit{cost})$
        \ELSE
            \STATE $p[i] \gets \min(p[i], \; p[t] + \textit{cost} + c_{a_i,0})$
        \ENDIF
    \ENDFOR
\ENDFOR
\RETURN $p[n]$

\end{algorithmic}
\end{algorithm}

\subsection{Model Architecture}

The route-first module is parameterized by a neural network consisting of an transformer-based encoder-decoder architecture augmented with history-enhanced context representation.
The architecture is designed to handle various CVRP variants within a unified network architecture.

\textbf{Encoder: } The encoder maps the raw node features into a latent embedding space.  
Let the node attributes for the $i$-th customer be
$\mathbf{x}_i = [x_i, y_i, q_i, r_i, s_i, d_i] \in \mathbb{R}^6$.
For the depot node, the attribute is represented as $\mathbf{x}_0 = [x_0, y_0, \mathrm{O}, \mathrm{B}, \mathrm{L}, \mathrm{TW}, l] \in \mathbb{R}^7$.

The node features are projected via learnable linear layers:
\[
\mathbf{h}_i^{(0)} = \begin{cases} 
\text{MLP}_\text{customer}(\mathbf{x}_i), & i \neq 0 \\
\text{MLP}_\text{depot}(\mathbf{x}_0), & i = 0
\end{cases}.
\]

The sequence of initial embeddings $\mathbf{H}^{(0)} = [\mathbf{h}_0^{(0)}, \dots, \mathbf{h}_n^{(0)}] \in \mathbb{R}^{(n+1) \times d}$ is then processed by a stack of multiple Routefinder Transformer \cite{berto2024routefinder} blocks.  
Each block consists of:
\begin{enumerate}
    \item Multi-head self-attention:  \\ $\mathbf{\overline{H}}^{(i)} = \mathbf{H}^{(i-1)} + \text{MHA}(\text{RMSNorm}(\mathbf{H}^{(i-1)}))$,
    \item Gated feedforward network: \\ $\mathbf{H}^{(i)} = \mathbf{\overline{H}}^{(i)} + \text{GatedMLP}(\text{RMSNorm}(\mathbf{\overline{H}}^{(i)}))$.
\end{enumerate}
A final RMS normalization is applied to produce node embeddings $\mathbf{H}^{(N)} \in \mathbb{R}^{(n+1) \times d}$ (Hereafter, $(N)$ is omitted for notational simplicity), and a graph-level embedding $\mathbf{g}$ is computed by mean pooling of $\mathbf{H}$.

\textbf{Decoder: } At each decoding step \(t\), the policy selects the next node \(a_t\) based on a context embedding \(\mathbf{c}_t \in \mathbb{R}^d\) computed from:
\[
\mathbf{c}_t = \text{LSTM} \big( [\mathbf{g}, \mathbf{h}_{0}, \mathbf{h}_{\text{current}}, \mathbf{d}_t] \big),
\]
where $\mathbf{h}_{0}$ and $\mathbf{h}_{\text{current}}$ are embeddings of the depot and current nodes, and \(\mathbf{d}_t = \text{Linear}([n_0, n_1])\) encodes the cumulative delivery/pickup loads.  
This history-enhanced context enables the agent to maintain a hidden state capturing historical information, mitigating partial observability induced by the knowledge-embedded RFCS framework.

A multi-head attention mechanism is applied between \(\mathbf{c}_t\) and the node embeddings \(\mathbf{H}\):
\[
\mathbf{y}_t = \text{MHA}(\mathbf{c}_t, \mathbf{H}, \mathbf{H}, \text{mask} = \mathbf{m}_t),
\]
where \(\mathbf{m}_t\) is the action mask. The resulting attention output is projected back into the embedding space, producing a compatibility score for each candidate node:
\[
\pi_\theta(a_t \mid s_t) \propto \exp(\alpha \cdot \tanh(\mathbf{y}_t^\top \mathbf{H})),
\]
with infeasible actions masked by \(-\infty\). Here, $\alpha = 10$ is a temperature scaling factor controlling exploration.

This architecture ensures that the same network can process all CVRP variants without variant-specific modifications.  
The decoder outputs a probability distribution over feasible next nodes, from which the action is sampled or greedily selected.
The network is trained via REINFORCE \cite{williams2025} accroding to equation (\ref{eq:reinforce_objective}).

\section{Experiments}

We empirically validate the effectiveness of the proposed knowledge-embedded framework in this section.
Our experimental design examines three key aspects: solution quality, generalization ability, and ablation analysis—corresponding to the following research questions:

\begin{enumerate}
    \item[RQ1] Does the proposed knowledge-enhanced framework improve solution quality?
    \item[RQ2] How well does the proposed framework generalize across different problem instances?
    \item[RQ3] How does the presence or absence of historical information in the network architecture and the learning process influence the framework's performance?
\end{enumerate}

\subsection{Experiment Settings}

Following the data generation process in \cite{berto2024routefinder}, depot and customer coordinates are uniformly sampled within the unit square.
For training instances with 50 and 100 customers, the vehicle capacities $Q$ are set to 40 and 50, respectively, while customer demands are uniformly sampled from $\{1, 2, \cdots, 9\}$.
In the backhaul scenario, 20\% of the customers are randomly assigned negative demands to simulate pickup requests.
For the time-window constraints, the service duration of each customer is uniformly sampled from $[0.15, 0.18]$.
The length of the time window is drawn uniformly from $[0.18, 0.2]$, and both the starting and ending times are generated by applying a random offset uniformly sampled from the feasible range.
The distance limit is uniformly drawn from $[2 D_{max}, 3]$, where $D_{max}$ denotes the maximum Euclidean distance between the depot and the farthest customer node.

During training, all problem attributes are randomly generated on the fly, and the constraint flags are uniformly sampled across different variants.
For each training batch, we fix one particular constraint configuration so that all instances in the batch share the same problem setting.
For each problem setting, the model is trained for 300 epochs, with 1,280,000 samples per epoch and a batch size of 512.
To stabilize training and improve final convergence, the learning rate is progressively decreased toward the end of training.
The framework demonstrates better performance without using POMO-style multi-start \cite{kwon2020pomo}.
During training, we simply perform eight stochastic policy rollouts per instance and take their average cost as the baseline.

We evaluate our framework using the open-source dataset from \cite{berto2024routefinder}.
We do not use multi-start sampling; instead, for each instance, we perform 8-fold data augmentation and generate as many trajectories as there are nodes, selecting the best among them.
This approach ensures consistent comparison data and produces an identical number of trajectories for all methods.
The code can be found at \url{https://github.com/wenwenla/ijcai26-cvrp-solver}.

\begin{table*}[t]
\centering
\caption{\textbf{Evaluation Results for RQ1}. The results are averaged over 1000 instances per variant. Lower values indicate better results. Bold numbers denote the best results among learning-based methods, while italic numbers represent the best-known results. }
\label{tab:tab1}
\resizebox{\textwidth}{!}{%
\begin{tabular*}{\textwidth}{@{\extracolsep{\fill}} c|cccccccc}
\toprule

n=50 & CVRP & OVRP & VRPB & VRPL & VRPTW & OVRPB & OVRPL & OVRPTW \\
\midrule
PyVRP & \textit{10.372} & \textit{6.507} & \textit{9.687} & \textit{10.587} & \textit{16.031} & \textit{6.898} & \textit{6.507} & \textit{10.510} \\
Or-Tools & 10.572 & 6.553 & 9.802 & 10.776 & 16.089 & 6.928 & 6.552 & 10.519 \\
MTPOMO & 10.518 & 6.718 & 10.033 & 10.775 & 16.410 & 7.108 & 6.719 & 10.668\\
MvMOE & 10.501 & 6.702 & 10.005 & 10.751 & 16.404 & 7.089 & 6.707 & 10.669\\
RouteFinder & 10.499 & 6.684 & 9.977 & \textbf{10.737} & 16.364 & 7.071 & 6.686 & 10.652\\
RFCS (ours) & \textbf{10.484} & \textbf{6.600} & \textbf{9.873} & 10.750 & \textbf{16.228} & \textbf{6.986} & \textbf{6.601} & \textbf{10.571} \\
\midrule

n=50 & VRPBL & VRPBTW & VRPLTW & OVRPBL & OVRPBTW & OVRPLTW & VRPBLTW & OVRPBLTW \\
\midrule
PyVRP & \textit{10.186} & \textit{18.292} & \textit{16.356} & \textit{6.899} & \textit{11.669} & \textit{10.510} & \textit{18.589} & \textit{11.668} \\
Or-Tools  & 10.331 & 18.366 & 16.441 & 6.927 & 11.682 & 10.519 & 18.694 & 11.681 \\
MTPOMO & 10.672 & 18.639 & 16.824 & 7.112 & 11.814 & 10.670 & 18.990 & 11.817\\
MvMOE &  10.637 & 18.640 & 16.811 & 7.098 & 11.819 & 10.671 & 18.985 & 11.822\\
RouteFinder & 10.575 & 18.600 & 16.750 & 7.074 & 11.805 & 10.653 & 18.937 & 11.805 \\
RFCS (ours) & \textbf{10.475} & \textbf{18.464} & \textbf{16.650} & \textbf{6.987} & \textbf{11.724} & \textbf{10.571} & \textbf{18.849} & \textbf{11.723} \\
\midrule

n=100 & CVRP & OVRP & VRPB & VRPL & VRPTW & OVRPB & OVRPL & OVRPTW \\
\midrule
PyVRP & \textit{15.628} & \textit{9.725} & \textit{14.377} & \textit{15.766} & \textit{25.423} & \textit{10.335} & \textit{9.724} & \textit{16.926} \\
Or-Tools & 16.280 & 9.995 & 14.933 & 16.407 & 25.814 & 10.577 & 10.001 & 17.027\\
MTPOMO & 15.934 & 10.210 & 15.082 & 16.149 & 26.412 & 10.878 & 10.214 & 17.420\\
MvMOE & 15.888 & 10.177 & 15.023 & 16.099 & 26.389 & 10.840 & 10.184 & 17.416\\
RouteFinder & 15.857 & 10.121 & 14.942 & \textbf{16.051} & 26.235 & 10.772 & 10.120 & 17.327 \\
RFCS (ours) & \textbf{15.837} & \textbf{10.010} & \textbf{14.840} & 16.064 & \textbf{26.047} & \textbf{10.645} & \textbf{10.009} & \textbf{17.158} \\
\midrule

n=100 & VRPBL & VRPBTW & VRPLTW & OVRPBL & OVRPBTW & OVRPLTW & VRPBLTW & OVRPBLTW \\
\midrule
PyVRP & \textit{14.779} & \textit{29.467} & \textit{25.757} & \textit{10.335} & \textit{19.156} & \textit{16.926} & \textit{29.810} & \textit{19.156} \\
Or-Tools & 15.426 & 29.945 & 26.259 & 10.582 & 19.303 & 17.027 & 30.396 & 19.305 \\
MTPOMO & 15.712 & 30.437 & 26.891 & 10.884 & 19.635 & 17.420 & 30.898 & 19.637 \\
MvMOE & 15.640 & 30.436 & 26.868 & 10.847 & 19.638 & 17.419 & 30.892 & 19.641 \\
RouteFinder & 15.528 & 30.241 & 26.689 & 10.778 & 19.550 & 17.327 & 30.688 & 19.551 \\
RFCS (ours) & \textbf{15.438} & \textbf{30.058} & \textbf{26.580} & \textbf{10.648} & \textbf{19.381} & \textbf{17.159} & \textbf{30.577} & \textbf{19.381} \\
\bottomrule

\end{tabular*}%
}
\end{table*}

\subsection{Baselines}\label{sec:baselines}

We select the following methods as baselines for comparison, which are described in detail below.
These solvers are open-source, allowing reproducible comparisons.
We select two classical optimization algorithms, PyVRP \cite{Wouda_Lan_Kool_PyVRP_2024} and OR-Tools \cite{ortools_routing}, as non-learning baselines for comparison.
We also compare with three recent learning-based VRP solvers:  (1) MTPOMO \cite{liu2024multi}, a multi-task VRP solver based on POMO \cite{kwon2020pomo}; (2) MVMoE \cite{zhou2024mvmoe}, a multi-task VRP solver leveraging mixture-of-experts networks for unified handling of different VRP variants;  and
(3) RouteFinder \cite{berto2024routefinder}, a transformer-based foundation model designed for general VRPs.  

\subsection{Results for RQ1}\label{sec:rq1_result}

\begin{figure}[htb]
	\centering
	\includegraphics[width=0.8\linewidth]{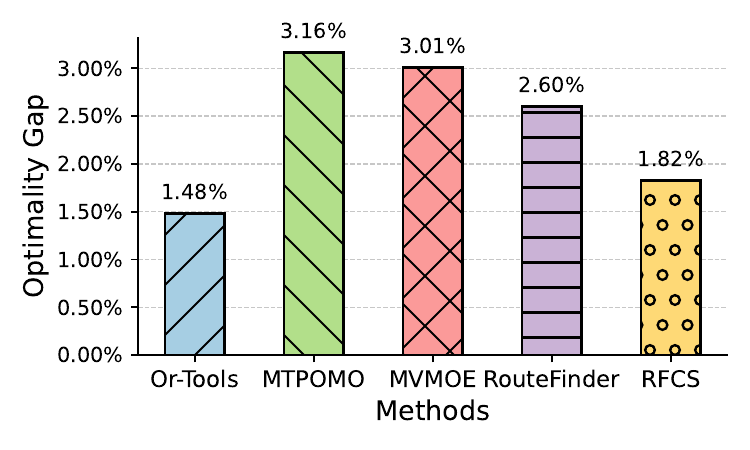}
     \vspace{-15 pt}
	\caption{The Average Optimality Gap.}
	\label{fig:opt_gap}
\end{figure}

Table~\ref{tab:tab1} presents the average costs over 1,000 unseen instances \cite{berto2024routefinder} for sixteen CVRP variants, including the atomic problems—CVRP, OVRP, VRPB, VRPL, and VRPTW—as well as their various combinations.
The knowledge-embedded framework outperforms all learning-based methods across all variants, except for the VRPL variant, where it performs slightly below RouteFinder.
The average optimality gap are shown in Figure \ref{fig:opt_gap}.
RFCS achieves a marked improvement by reducing the average optimality gap from 2.60\% to 1.82\% on average, highlighting the stronger solution quality achieved by our framework.
Since the proposed framework shares the same neural network forward-propagation cost as other learning-based methods, and the dynamic programming-based splitting algorithm has a time complexity of $O(n^2)$, its computational overhead is negligible compared with neural network inference.
Therefore, the overall solving time of our framework is comparable to that of the baseline learning-based methods.
While classic approaches often yield better solution quality, they usually demand tens of minutes to complete a computation, in contrast to learning-based methods, which require only a few seconds.
For more discussions regarding runtime and scalability, please refer to the appendix.

\subsection{Results for RQ2}\label{sec:rq2_result}

We evaluate the zero-shot transfer capability of RFCS across different capacity distributions encountered during training.
During training, for scenarios with $n=50$, the vehicle capacity was set to 40. 
We directly evaluate the model obtained in RQ1 on instances with varying vehicle capacities to examine its performance changes. 
The target distribution of vehicle capacities ranges from 50 to 90, and the experimental results of zero-shot transfer evaluation are presented in Table \ref{tab:t_cap}.

\begin{table}[!htbp]
    \centering
    \caption{Evaluation of the Knowledge-embedded Framework Zero-Shot Transfer to Unseen Vehicle Capacity Distributions.}\label{tab:t_cap}
    \begin{tabular}{c|ccccc}
    \toprule
     Capacity & 50 & 60 & 70 & 80 & 90   \\
    \midrule
    MTPOMO & 9.255 & 8.482 & 7.950 & 7.553 & 7.269 \\
    MvMOE & 9.237 & 8.456 & 7.915 & 7.511 & 7.206 \\
    RouteFinder & 9.238 & 8.453 & 7.908 & 7.487 & 7.180 \\
    RFCS (ours) & \textbf{9.214} & \textbf{8.433} & \textbf{7.882} & \textbf{7.469} & \textbf{7.154} \\
    \bottomrule    
    \end{tabular}
\end{table}

The experimental results indicate that when the vehicle capacity distributions differ between training and evaluation, the knowledge-embedded framework exhibits superior zero-shot transfer performance compared to existing learning-based methods.

A commonly studied CVRP variant introduces mixed linehaul and backhaul demands \cite{zhou2024mvmoe}, allowing vehicles to alternate between delivery and pickup operations instead of strictly finishing deliveries before pickups.
In this case, the key constraint is that the vehicle load must not violate the capacity limit at any time.
As described in section \ref{sec:dp_rewards}, we only need to modify the Algorithm \ref{alg:msdp} to adapt the solver to this scenario.
The zero-shot transfer results to mixed demands are shown in Table~\ref{tab:b2mp}.
The results show that our method outperforms the baselines when transferring from VRPB and OVRPB to the mixed constraint scenario, achieves comparable performance for VRPBL, and exhibits a notable performance drop for VRPBTW.
While the framework does not surpass the baselines across every test case, the findings highlight its inherent flexibility and adaptability.

\begin{table}[!htbp]
    \centering
    \caption{Evaluation of the Knowledge-embedded Framework Zero-Shot Transfer to Unseen Mixed Backhaul Constraints.}\label{tab:b2mp}
    \begin{tabular}{c|cccc}
    \toprule
     Source & VRPB & OVRPB & VRPBL & VRPBTW \\
    \midrule
    MTPOMO & 9.900 & 7.002 & 10.284 & 17.257 \\
    MvMOE & 9.857 & 6.990 & 10.241 & \textbf{17.103} \\
    RouteFinder & 9.879 & 6.948 & \textbf{10.239} & 17.298 \\
    RFCS (ours) & \textbf{9.534} & \textbf{6.721} & 10.301 & 18.194\\
    \bottomrule    
    \end{tabular}
\end{table}

\subsection{Results for RQ3}\label{sec:rq3_result}

We conduct an ablation study to assess the contribution of the history-enhanced context in the knowledge-embedded framework. 
The results are summarized in Table~\ref{tab:ablation_study}.
For the variant without historical information, we adopt an identical training setup, except that the LSTM module in the decoder responsible for handling historical information is replaced with a fully connected layer.
The results of this ablation study highlight the necessity of incorporating historical information.
While the LSTM is not the most advanced architecture for modeling history, it was chosen for its ability to incrementally maintain past information.
Further improvements in processing historical information by introduce more advanced network represent an interesting direction for future research.

\begin{table}[!htbp]
    \centering
    \caption{Effect of Historical Information on Solution Quality: Ablation Results for 50-node CVRP Variants. ``Gap'' denotes the optimality gap relative to the best-known solutions.}\label{tab:ablation_study}
    \begin{tabular}{c|ccccc}
    \toprule
     &  CVRP & OVRP & VRPB & VRPL & VRPTW \\
    \midrule
    RFCS & \textbf{10.484} & \textbf{6.600} & \textbf{9.873} & \textbf{10.750} & \textbf{16.228}\\
    Gap & 1.08\% & 1.43\% & 1.92\% & 1.54\% & 1.23\% \\
    \midrule
    - History & 10.720 & 6.667 & 10.035 & 10.990 & 16.271 \\
    Gap & 3.36\% & 2.46\% & 3.59\% & 3.81\% & 1.50\% \\
    \bottomrule    
    \end{tabular}
\end{table}

Another interesting ablation concerns whether incorporating the dynamic programming-based reward  during training is necessary, and whether the route-first component actually learns to cooperate with the dynamic programming  to produce higher-quality solutions.
To address this question, we examine the results obtained by directly applying the cluster-second dynamic programming module to optimize the solutions for CVRP generated by RouteFinder~\cite{berto2024routefinder}.
The corresponding results are shown in Table~\ref{tab:ablation_removedp}.
From the table, we observe that applying the cluster-second optimization improves the solutions generated by the learning-based methods.
However, the improvement is marginal, and the resulting performance remains inferior to that of RFCS trained with the dynamic programming-based reward.
This not only highlights the optimality improvement provided by the DP procedure, but also demonstrates that incorporating the DP-based reward during training is necessary.

\begin{table}[!htbp]
    \centering
    \caption{Ablation Study on the Necessity of Dynamic Programming-Based Rewards in RFCS Training.}\label{tab:ablation_removedp}
    \begin{tabular}{c|ccc}
    \toprule
     &  Original & AfterDP & Reduced \\
    \midrule
    MTPOMO & 10.518 & 10.516 & 0.02 \% \\
    MvMOE & 10.501 & 10.499 & 0.02 \% \\
    RouteFinder & 10.499 & 10.498 & 0.01 \% \\
    RFCS (ours) & - & \textbf{10.484} & - \\
    \bottomrule    
    \end{tabular} 
\end{table}

\section{Conclusion}

In this paper, we propose a unified framework for a broad range of CVRP variants. 
Inspired by the route-first cluster-second (RFCS) heuristic, the framework integrates RL with dynamic programming-based algorithmic design knowledge to solve CVRPs with various additional constraints.
Specifically, the route-first component leverages RL to generate customer visitation sequences.
The cluster-second component employs dynamic programming to optimally convert a given sequence into a feasible solution for the original problem, and provides the resulting cost as a reward to guide RL.
The proposed framework not only achieves strong performance in terms of solution quality and transferability but also offers a flexible and unified approach to handling diverse constraints.
In future work, we plan to incorporate more advanced network architectures with this framework to achieve even better results.
We also plan to extend this approach to other domains beyond routing problems.

\newpage

\appendix



\section*{Acknowledgments}

This work was supported by the NSFC Major Research Plan Key Program under Grant No. 92582204, and the Collaborative Innovation Center of Novel Software Technology and Industrialization.

\section{Scalability and Inference Time}

To further assess scalability, we conducted additional zero-shot experiments by training the model on \textbf{50-node} instances and directly transferring it to \textbf{500-node and 1000-node} CVRP without fine-tuning.
The results are reported in Table \ref{tab:scale_large}. 
For POMO-based methods, the number of starts typically scales with the number of nodes. 
However, due to GPU memory limitations on large-scale instances, we evaluate these methods using $1$ and $100$ starts.
As shown in Table \ref{tab:scale_large}, our method achieves consistently better solution quality on both 500-node and 1000-node instances. 
These results suggest that the proposed framework maintains stable performance when scaling to larger problem sizes and demonstrates better zero-shot generalization ability. 
The time is for solving 100 instances.

\begin{table}[!h]
\centering
\small
\setlength{\tabcolsep}{4pt}
\begin{tabular}{lcccc}
\hline
& \multicolumn{2}{c}{n=500} & \multicolumn{2}{c}{n=1000} \\
\cline{2-5}
Method & Cost $\downarrow$ & Time (s) $\downarrow$ & Cost $\downarrow$ & Time (s) $\downarrow$ \\
\hline
m1 (s$_1$)       & 59.278 & 1.967 & 127.206 & 4.478  \\
m1 (s$_{100}$)     & 53.646 & 7.769 & 113.193 & 30.939 \\

m2 (s$_1$)       & 52.834 & 1.758 & 114.655 & 3.887  \\
m2 (s$_{100}$)     & 51.295 & 6.885 & 110.193 & 26.670 \\

m3 (s$_1$)       & 78.523 & 2.222 & 182.017 & 4.976  \\
m3 (s$_{100}$)     & 75.465 & 7.590 & 173.143 & 30.293 \\

RFCS (s$_1$)     & \textbf{45.274} & 1.650 & \textbf{72.891} & 4.850 \\
\hline
\end{tabular}
\caption{Zero-shot generalization performance on large-scale CVRP instances. 
$s_n$ denotes the number of POMO starts. All methods are evaluated with greedy rollout. 
m1: RouteFinder, m2: MTPOMO, m3: MvMoE. 
Costs are averaged over 100 random instances.}
\label{tab:scale_large}
\end{table}

To access inference time, from a \textbf{theoretical} standpoint, the proposed framework \textbf{does not increase the asymptotic time complexity} compared to existing end-to-end learning-based VRP solvers. 
Specifically, the DP-based clustering stage has complexity $O(n^2)$, which is of the same asymptotic order as the attention computation in transformer-based methods.
From a \textbf{practical} perspective, empirical results on large-scale instances are provided in Table \ref{tab:scale_large}.
The results are obtained on a Linux server with an RTX5090 and 8470Q with 208 cores. 
The DP procedure is implemented in C++ and integrated into the training and evaluation pipeline via pybind11.

\section{Motivation of Learning}

We apply LKH3 to solve the same 1,000 TSP instances with $n=50$. 
For each instance, the resulting tour is circularly permuted 50 times so each node serves as the start, and the tours are fed into the DP-based clustering module. 
Results: $\textbf{11.312}$ vs $\textbf{10.484}$.
These results suggest that even high-quality TSP tours do not necessarily lead to strong CVRP solutions after clustering, highlighting the importance of learning a task-aware sequence construction policy. 

\section{Additional Experiment Details}

Since RouteFinder \cite{berto2024routefinder} released the benchmark dataset and pretrained models, we directly report the results of non-learning solvers (PyVRP and OR-Tools) from the original paper for fair comparison, where the time limits are \textbf{10s and 20s} per instance for n=50 and n=100, respectively.
Regarding training efficiency, training time depends on hardware and implementation and may vary significantly across works. 
For example, for n=50, MvMoE reports training times ranging from 38–90 hours for different architectures, while RouteFinder reports 9–24 hours using A100 GPUs. 
In comparison, our model is trained on two RTX 5090 GPUs and requires \textbf{16.17 hours} for n=50, which is comparable to existing learning-based approaches.

\section{Discussion about Historical Information}

The history-modeling component is modular and can be readily replaced by other \textbf{RNN} architectures. 
We chose LSTM as it maintains hidden states \textbf{incrementally}, offering better efficiency and memory scalability than Transformers that attend to the full sequence at each step. 
Exploring advanced alternatives is a promising future direction. 
For VRPBTW, the history ablation ($18.464 \to 18.489$) also confirms the module's effectiveness under mixed constraints.

\bibliographystyle{named}
\bibliography{ijcai26}

\end{document}